\documentclass[runningheads]{llncs}
\usepackage[utf8]{inputenc}
\usepackage{color}
\usepackage{subfig}
\usepackage{graphicx}
\usepackage{tabularx}
\usepackage{multirow}
\usepackage{booktabs}
\usepackage{amsmath}

\newcolumntype{L}[1]{>{\raggedright\let\newline\\\arraybackslash\hspace{0pt}}m{#1}}
\newcolumntype{C}[1]{>{\centering\let\newline\\\arraybackslash\hspace{0pt}}m{#1}}
\newcolumntype{R}[1]{>{\raggedleft\let\newline\\\arraybackslash\hspace{0pt}}m{#1}}

\title{Quantifying and Leveraging Classification Uncertainty for Chest Radiograph Assessment}

\ifx\anonymize\undefined
\author{Florin C. Ghesu\inst{1}\and Bogdan Georgescu\inst{1}\and Eli Gibson\inst{1}\and Sebastian Guendel\inst{1}\and\\Mannudeep K. Kalra\inst{2,3}\and Ramandeep Singh\inst{2,3}\and Subba R. Digumarthy\inst{2,3}\and\\Sasa Grbic\inst{1}\and Dorin Comaniciu\inst{1}}

\authorrunning{F. Ghesu et al.}

\institute{Digital Technology and Innovation, Siemens Healthineers, Princeton, NJ, USA\and
Department of Radiology, Massachusetts General Hospital, Boston, MA, USA\and
Harvard Medical School, Boston, MA, USA
\email{florin.ghesu@siemens-healthineers.com}
}
\index{Ghesu, Florin C.}
\index{Georgescu, Bogdan}
\index{Gibson, Eli}
\index{Guendel, Sebastian}
\index{Grbic, Sasa}
\index{Comaniciu, Dorin}

\else
\author{*}
\authorrunning{*}
\institute{*}
\fi

\begin{document}

\maketitle

\begin{abstract}
The interpretation of chest radiographs is an essential task for the detection of thoracic diseases and abnormalities. However, it is a challenging problem with high inter-rater variability and inherent ambiguity due to inconclusive evidence in the data, limited data quality or subjective definitions of disease appearance. Current deep learning solutions for chest radiograph abnormality classification are typically limited to providing probabilistic predictions, relying on the capacity of learning models to adapt to the high degree of label noise and become robust to the enumerated causal factors. In practice, however, this leads to overconfident systems with poor generalization on unseen data. To account for this, we propose an automatic system that learns not only the probabilistic estimate on the presence of an abnormality, but also an explicit uncertainty measure which captures the confidence of the system in the predicted output. We argue that explicitly learning the classification uncertainty as an orthogonal measure to the predicted output, is essential to account for the inherent variability characteristic of this data. Experiments were conducted on two datasets of chest radiographs of over 85,000 patients. Sample rejection based on the predicted uncertainty can significantly improve the ROC-AUC, e.g., by 8\% to 0.91 with an expected rejection rate of under 25\%. Eliminating training samples using uncertainty-driven bootstrapping, enables a significant increase in robustness and accuracy. In addition, we present a multi-reader study showing that the predictive uncertainty is indicative of reader errors.
\end{abstract}

\section{Introduction}

The interpretation of chest radiographs is an essential task in the practice of a radiologist, enabling the early detection of thoracic diseases~\cite{Rajpurkar2018,Wang2017}. To accelerate and improve the assessment of the continuously increasing number of radiographs, several deep learning solutions have been recently proposed for the automatic classification of radiographic findings~\cite{Wang2017,Guendel2018,Yao2018}. Due to large variations in image quality or subjective definitions of disease appearance, there is a large inter-rate variability which leads to a high degree of label noise~\cite{Rajpurkar2018}. Modeling this variability when designing an automatic system for assessing this type of data is essential; an aspect which was not considered in previous work.

Using principles of information theory and subjective logic~\cite{Josang2016} based on the Dempster-Shafer framework for modeling of evidence~\cite{Dempster1968}, we present a method for training a system that generates both an image-level label and a classification uncertainty measure. We evaluate this system for classification of abnormalities on chest radiographs. The main contributions of this paper include:
\begin{enumerate}
    \item describing a system for jointly learning classification probabilities and classification uncertainty in a parametric model;
    \item proposing uncertainty-driven bootstrapping as a means to filter training samples with highest predictive uncertainty to improve robustness and accuracy;
    \item comparing methods for generating stochastic classifications to model classification uncertainty; 
    \item presenting an application of this system to identify cases with uncertain classification, yielding more accurate classification on the remaining cases;
    \item showing that the uncertainty measure can distinguish radiographs with correct and incorrect labels according to a multi-radiologist-consensus study.
    
\end{enumerate}

\section{Background and Motivation}
\subsection{Machine Learning for the Assessment of Chest Radiographs}
The open access to the ChestX-Ray8 dataset~\cite{Wang2017} of chest radiographs has led to a series of recent publications that propose machine learning based systems for disease classification. With this dataset, Wang et al.~\cite{Wang2017} also report a first performance baseline of a deep neural network at an average area under receiver operating characteristic curve (ROC-AUC) of 0.75. These results have been further improved by using multi-scale image analysis~\cite{Yao2018}, or by actively focusing the attention of the network on the most relevant sub-regions of the lungs~\cite{Guan2018}. State-of-the-art results on the official split of the ChestX-Ray8 dataset are reported in~\cite{Guendel2018} (avg. ROC-AUC of 0.81), using a location-aware dense neural network. In light of these contributions, a recent study compares the performance of such an AI system and 9 practicing radiologists~\cite{Rajpurkar2018}. While the study indicates that the system can surpass human performance, it also highlights the high variability among different expert radiologists for the reading of chest radiographs. The reported average specificity of the readers is very high (over 95\%), with an average sensitivity of 50\%$\,\pm$\,8\%. With such a large inter-rater variability, one may ask: How can real 'ground truth' data be obtained? Does the label noise affect the training? Current solutions do not consider this variability, which leads to models with overconfident predictions and limited generalization.\medskip\\
\textbf{Principles of Uncertainty Estimation:}\, One way to handle this challenge is to explicitly estimate the classification uncertainty from the data. Recent methods for uncertainty estimation in the context of deep learning rely on Bayesian estimation theory~\cite{Molchanov2017} or ensembles~\cite{Laks2017} and demonstrate increased robustness to out-of-distribution data. However, these approaches come with significant computational limitations; associated with the high complexity of sampling parameter spaces of deep models for Bayesian risk estimation; or associated with the challenge of managing ensembles of deep models. Sensoy et al.~\cite{Sensoy2018} propose an efficient alternative based on the theory of subjective logic~\cite{Josang2016}, training a deep neural network to estimate the sample uncertainty based on observed data.

\section{Proposed Method}

\begin{figure*}[t]
\centering
\subfloat[Confident negative]{
\includegraphics[height=2.8cm]{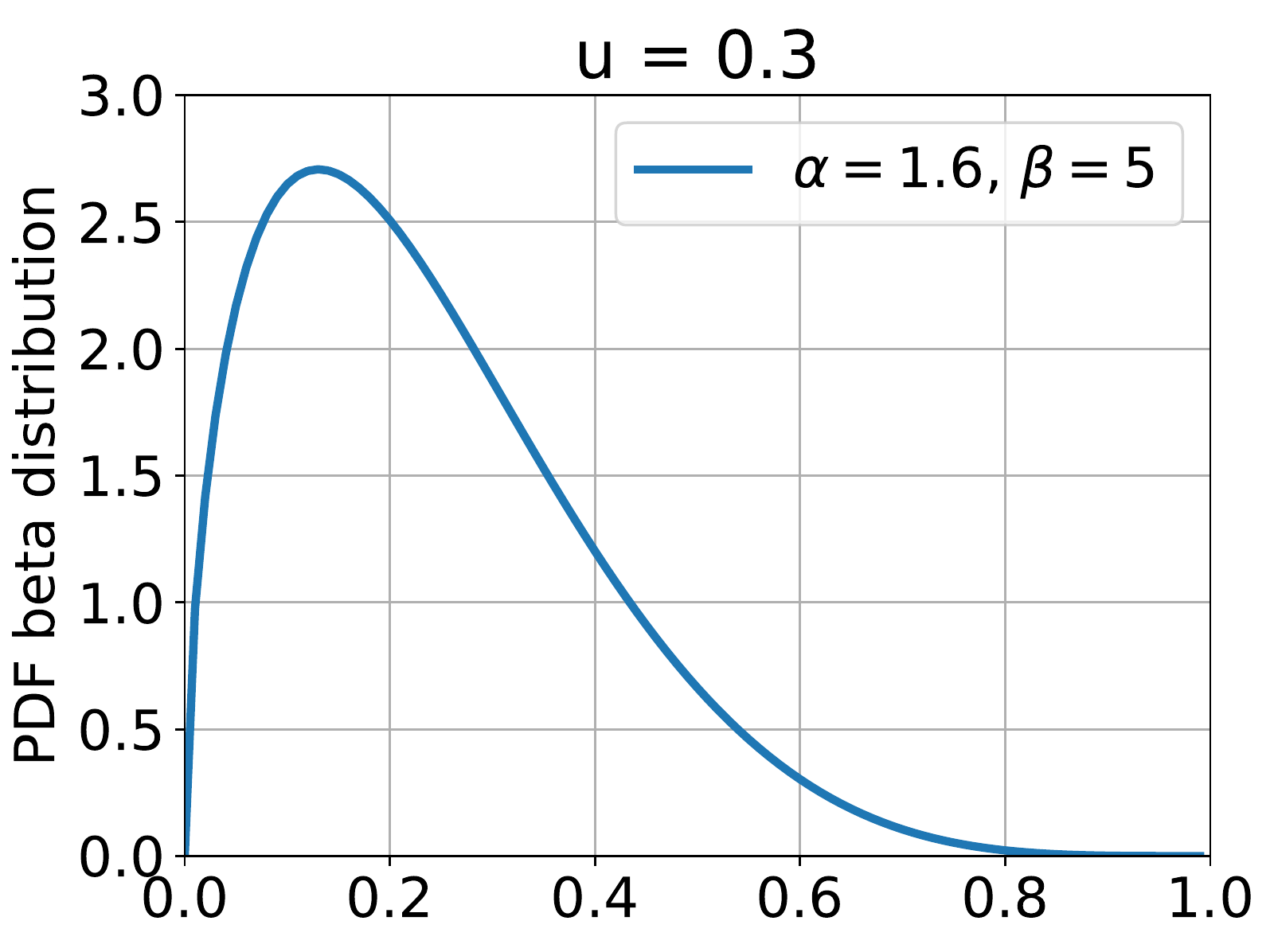}
}
\subfloat[Confident positive]{
\includegraphics[height=2.8cm]{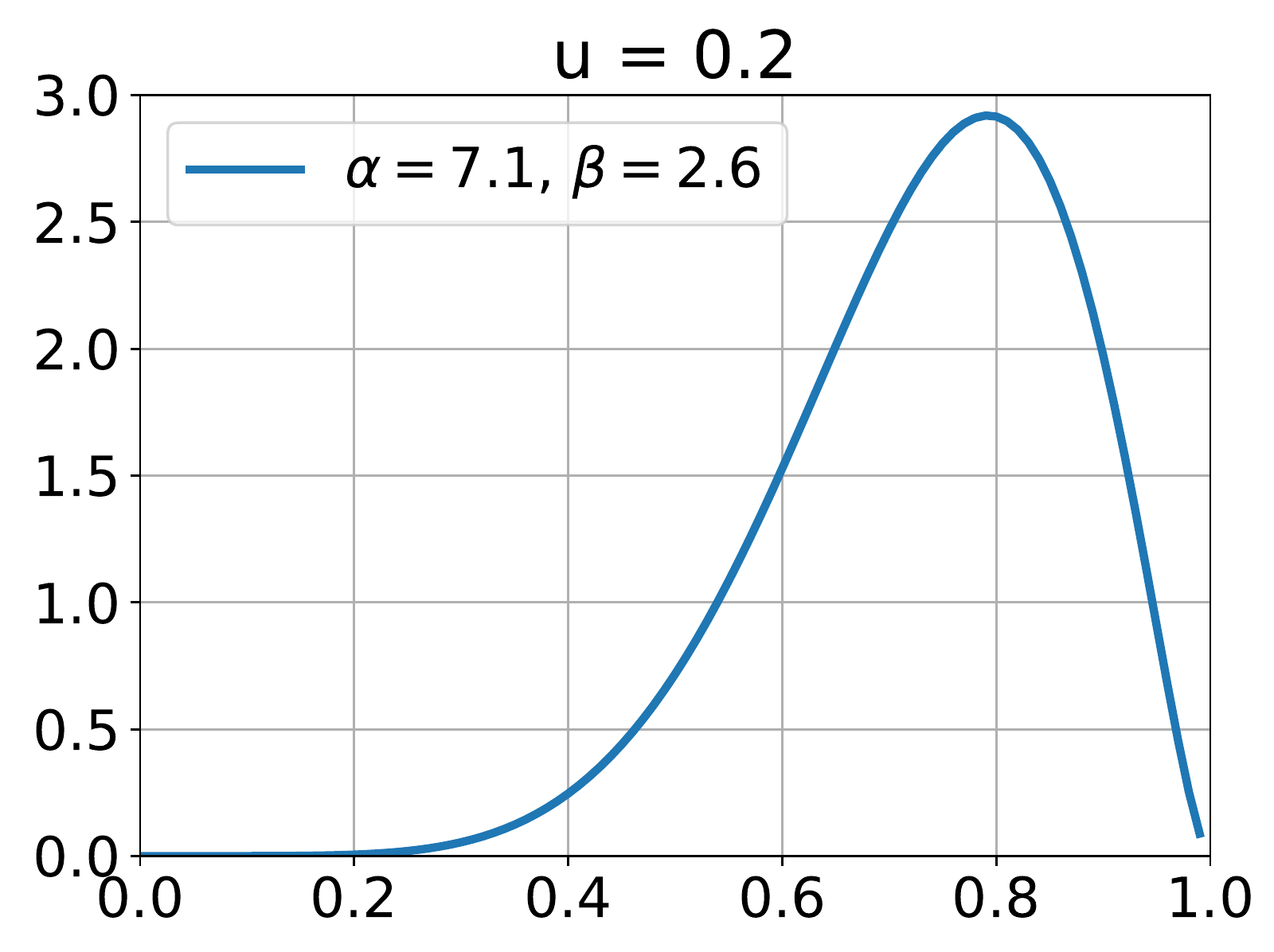}
}
\subfloat[High uncertainty]{
\includegraphics[height=2.8cm]{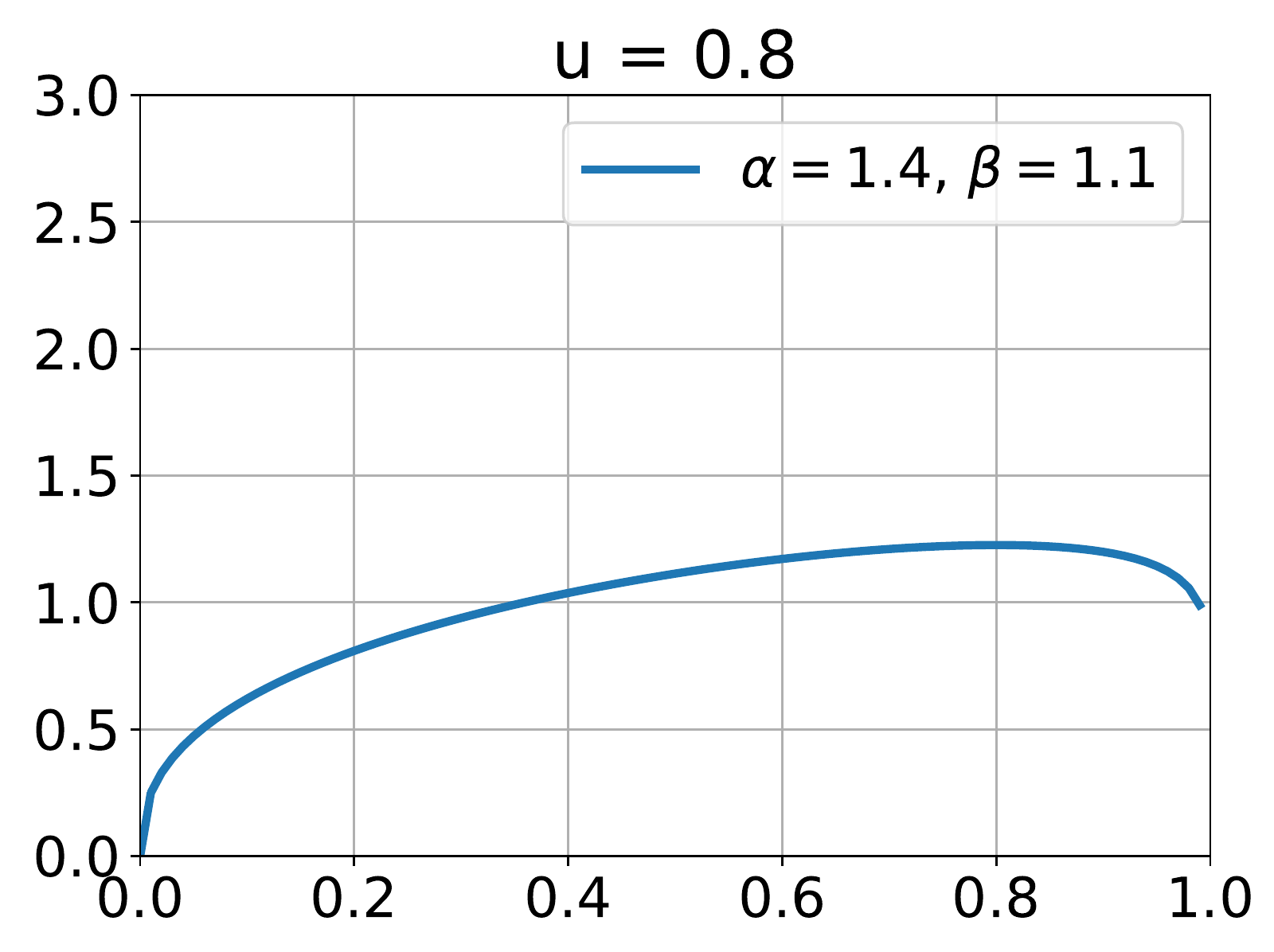}
}
\qquad
\caption{Probability density function of the beta distribution: example parameters ($\alpha, \beta$) modeling confident and uncertain predictions.\label{fig:beta}}
\end{figure*}

Following the work of Sensoy et al.~\cite{Sensoy2018} based on the Dempster-Shafer theory of evidence~\cite{Dempster1968}, we apply principles of subjective logic~\cite{Josang2016} to derive a binary classification model that can support the joint estimation of per-class probabilities ($\hat{p}_+; \hat{p}_-$) and predictive uncertainty $\hat{u}$. In this context, a decisional framework is defined through the assignment of so called belief masses from evidence collected from observed data to individual attributes, e.g., membership to a class~\cite{Dempster1968,Josang2016}. Let us denote $b^+$ and $b^-$ the belief values for the positive and negative class, respectively. The uncertainty mass $u$ is defined as: $u = 1 - b^+ - b^-$, where $b^+ = e^+/E$ and $b^- = e^-/E$ with $e^+; e^- \ge 0$ denoting the per-class collected evidence and total evidence $E = e^+ + e^- + 2$. For binary classification, we propose to model the distribution of such evidence values using the beta distribution, defined by two parameters $\alpha$ and $\beta$ as: $f(x;\alpha,\beta) = \frac{\Gamma(\alpha+\beta)}{\Gamma(\alpha)\Gamma(\beta)}x^{\alpha - 1}(1 - x)^{\beta - 1}$, where $\Gamma$ denotes the gamma function and $\alpha, \beta > 1$ with $\alpha = e^+ + 1$ and $\beta = e^- + 1$. In this context, the per-class probabilities can be derived as $p^+ = \alpha/E$ and $p^- = \beta/E$. Figure~\ref{fig:beta} visualizes the beta distribution for different $\alpha, \beta$ values.

A training dataset is provided: $\mathcal{D} = \{\vec{I}_k,y_k\}_{k=1}^{N}$, composed of $N$ pairs of images $\vec{I}_k$ with class assignment $y_k\in\{0,1\}$. To estimate the per-class evidence values from the observed data, a deep neural network parametrized by $\vec{\theta}$ can be applied, with: $[e^+_k, e^-_k] = \mathcal{R}(\vec{I}_k;\vec{\theta})$, where $\mathcal{R}$ denotes the network response function. Using maximum likelihood estimation, one can learn the network parameters $\hat{\vec{\theta}}$ by optimizing the Bayes risk with a beta distributed prior:
\begin{equation}
    \mathcal{L}^{data}_k = \int \|y_k - p_k\|^2 \frac{\Gamma(\alpha+\beta)}{\Gamma(\alpha)\Gamma(\beta)}p_k^{\alpha - 1}(1 - p_k)^{\beta - 1} dp_k,
\label{eq:lossmse}
\end{equation}
where $k\in\{1,\ldots,N\}$ denotes the index of the training example from dataset~$\mathcal{D}$, $p_k$ the predicted probability on the training sample $k$, and $\mathcal{L}^{data}_k$ defines the goodness of fit. Using linearity properties of the expectation, Eq.~\ref{eq:lossmse} becomes:
\begin{equation}
    \mathcal{L}^{data}_k = (y_k - \hat{p}_k^{\,+})^2 + (1 - y_k - \hat{p}_k^{\,-})^2 + \frac{\hat{p}_k^{\,+}(1 - \hat{p}_k^{\,+}) + \hat{p}_k^{\,-}(1 - \hat{p}_k^{\,-})}{E_k + 1},
\end{equation}
where $\hat{p}_k^{\,+}$ and $\hat{p}_k^{\,-}$ denote the network's probabilistic prediction. The first two terms measure the goodness of fit, and the last term encodes the variance of the prediction~\cite{Sensoy2018}.

To ensure a high uncertainty value for data samples for which the gathered evidence is not conclusive for an accurate classification, an additional regularization term $\mathcal{L}^{reg}_k$ is added to the loss. Using information theory, this term is defined as the relative entropy, i.e., the Kullback-Leibler divergence, between the beta distributed prior term and the beta distribution with total uncertainty ($\alpha, \beta = 1$). In this way, cost deviations from the total uncertainty state, i.e., $u = 1$, which do not contribute to the data fit are accounted for~\cite{Sensoy2018}. With the additional term, the total cost becomes $\mathcal{L} = \sum_{k=1}^{N}\mathcal{L}_k$ with:
\begin{equation}
    \mathcal{L}_k = \mathcal{L}^{data}_k + \lambda\,\text{KL}\left (f(\hat{p}_k;\tilde{\alpha}_k,\tilde{\beta}_k)\|f(\hat{p}_k;\langle 1,1\rangle)\right),
\end{equation}
where $\lambda\in[0,1]$, $\hat{p}_k = \hat{p}_k^{\,+}$, with $(\tilde{\alpha}_k, \tilde{\beta}_k)=(1, \beta_k)$ for $y_k = 0$ and $(\tilde{\alpha}_k, \tilde{\beta}_k)=(\alpha_k, 1)$ for $y_k = 1$. Removing additive constants and using properties of the logarithm function, one can simplify the regularization term to the following:
\begin{equation}
    \mathcal{L}^{reg}_k = \log\frac{\Gamma(\tilde{\alpha}_k+\tilde{\beta}_k)}{\Gamma(\tilde{\alpha}_k)\Gamma(\tilde{\beta_k})} + \sum_{x\in\{\tilde{\alpha}_k, \tilde{\beta}_k\}} (x - 1)\left(\psi(x) - \psi(\tilde{\alpha}_k + \tilde{\beta}_k)\right),
\end{equation}
where $\psi$ denotes the digamma function and $k\in\{1,\ldots,N\}$. Using stochastic gradient descent, the total loss $\mathcal{L}$ is optimized on the training set~$\mathcal{D}$.\smallskip

\textbf{Sampling the Data Distribution:}\, An important requirement to ensure training stability and to learn a robust estimation of evidence values is an adequate sampling of the data distribution. We empirically found dropout~\cite{Srivastava2014} to be a simple and very effective strategy to address this problem. In practice, dropout emulates an ensemble model combination driven by the random deactivation of neurons. Alternatively, one may use an explicit ensemble of $M$ models $\{\vec{\theta}_k\}_{k=1}^{M}$, each trained independently. Following the principles of deep ensembles~\cite{Laks2017}, the per-class evidence can be computed from the ensemble estimates $\{e^{(k)}\}_{k=1}^{M}$ via averaging. In our work, we found dropout to be as effective as deep ensembles.\smallskip

\textbf{Uncertainty-driven Bootstrapping:}\, Given the predictive uncertainty measure $\hat{u}$, we propose a simple and effective algorithm for filtering the training set with the target of reducing label noise. A fraction of training samples with highest uncertainty are eliminated and the model is retrained on the remaining data. Instead of sample elimination, robust M-estimators may be applied, using a per-sample weight that is inversely proportional to the predicted uncertainty. The hypothesis is that by focusing the training on 'confident' labels, we increase the robustness of the classifier and improve its performance on unseen data.

\section{Experiments}

\textbf{Dataset and Setup:}\, The evaluation is based on two datasets, the ChestX-Ray8~\cite{Wang2017} and PLCO~\cite{PLCO}. Both datasets provide a series of AP/PA chest radiographs with binary labels on the presence of different radiological findings, e.g., granuloma, pleural effusion, or consolidation. The ChestX-Ray8 dataset contains 112,120 images from 30,805 patients, covering 14 findings extracted from radiological reports using natural language processing (NLP)~\cite{Wang2017}. In contrast, the PLCO dataset was constructed as part of a screening trial, containing 185,421 images from 56,071 patients and covering 12 different abnormalities.

For performance comparison, we selected location-aware dense networks~\cite{Guendel2018} as baseline. This method achieves state-of-the-art results on this problem, with a reported average ROC-AUC of \textbf{0.81} (significantly higher than that of competing methods: 0.75~\cite{Wang2017} and 0.77~\cite{Yao2018}) on the official split of the ChestX-Ray8 dataset and a ROC-AUC of \textbf{0.88} on the official split of the PLCO dataset. To evaluate our method, we identified testing subsets with higher confidence labels from multi-radiologist studies. For PLCO, we randomly selected 565 test images and had 2 board-certified expert radiologists read the images -- updating the labels to the majority vote of the 3 opinions (incl. the original label). For ChestX-Ray8, a subset of 689 test images was randomly selected and read by 4 board-certified radiologists. The final label was decided following a consensus discussion. For both datasets, the remaining data was split in 90\% training and 10\% validation. All images were down-sampled to $256\times 256$ using bilinear interpolation.\smallskip

\textbf{System Training:}\, We constructed our learning model from the DenseNet-121 architecture~\cite{Huang2017}. A dropout layer with a dropout rate of 0.5 was inserted after the last convolutional layer. We also investigated the benefits of using deep ensembles to improve the sampling ($M=5$ models trained on random subsets of 80\% of the training data; we refer to this with the keyword \textbf{[ens]}). A fully connected layer with ReLU activation units maps to the two outputs $\alpha$ and $\beta$. We used a systematic grid search to find the optimal configuration of training meta-parameters: learning rate ($10^{-4}$), regularization factor ($\lambda=1$; decayed to $0.1$ and $0.001$ after 1/3, respectively 2/3 of the epochs), training epochs (around 12, using an early stop strategy with a patience of 3 epochs) and a batch size of 128. The low number of epochs is explained by the large size of the dataset.\smallskip

\begin{figure}[t]
\centering
\includegraphics[height=3.75cm]{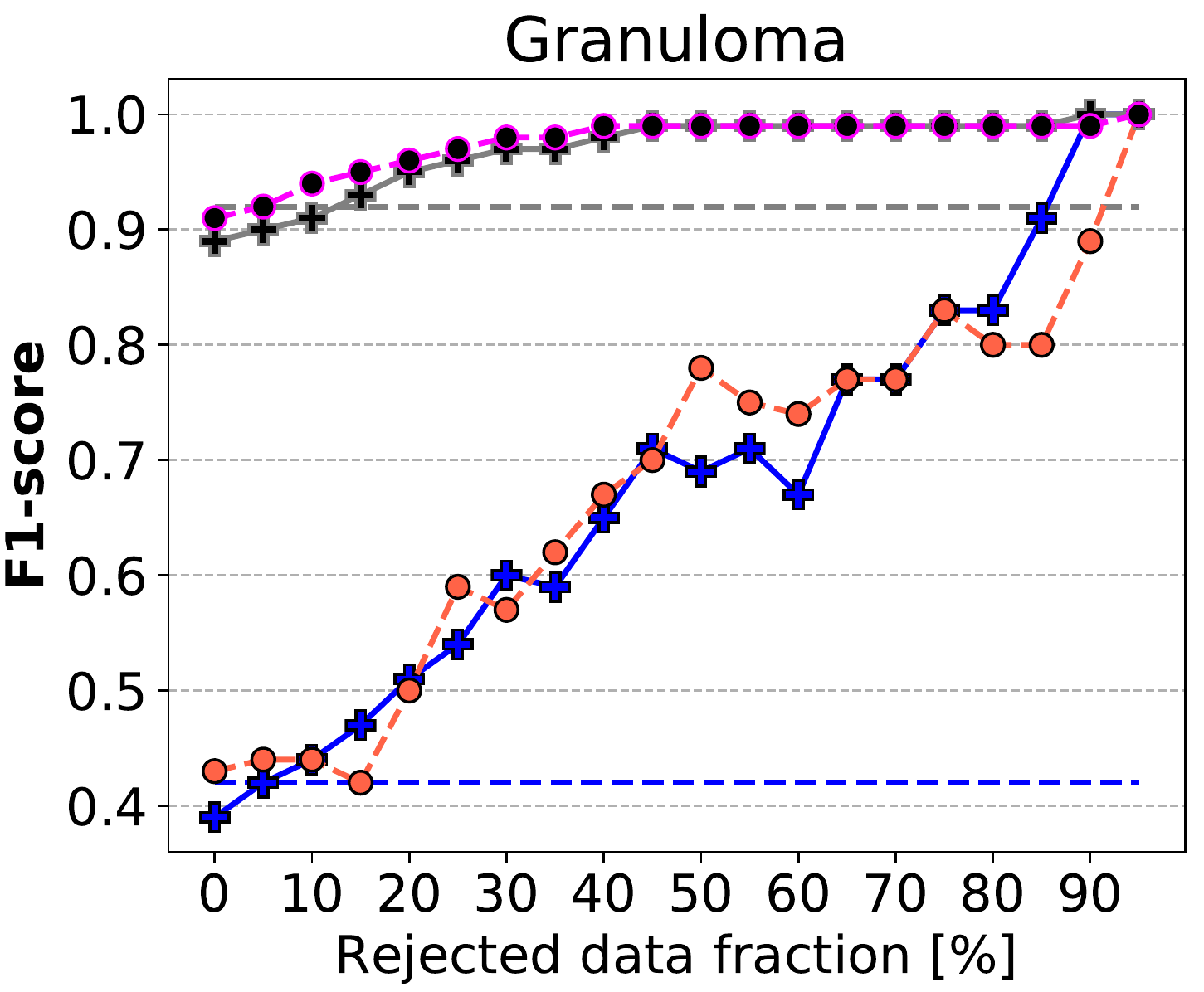}
\includegraphics[height=3.75cm]{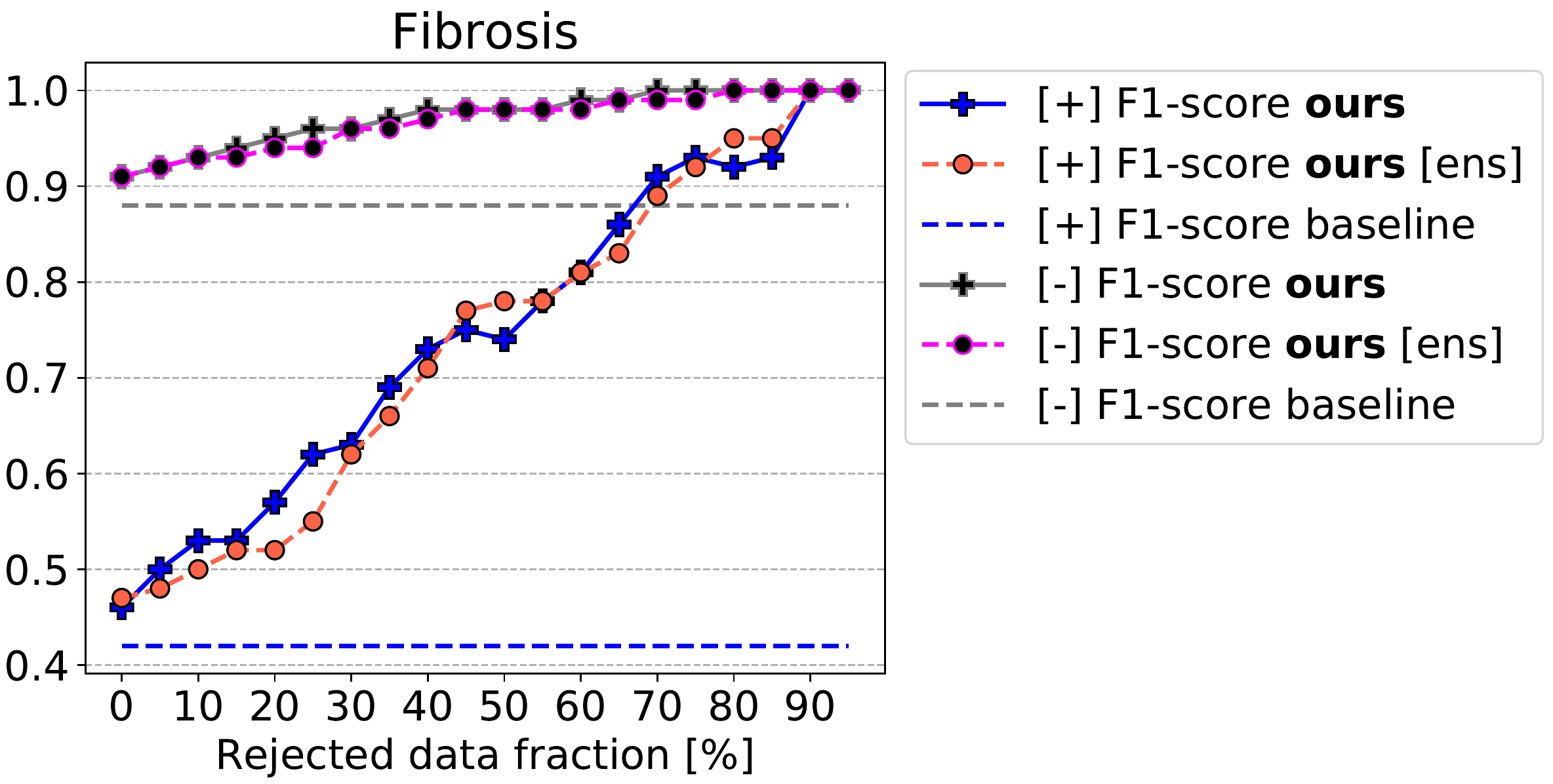}
\caption{Evolution of the F1-scores for the positive (+) and negative (--) classes relative to the sample rejection threshold - determined using the estimated uncertainty. We show the performance for granuloma and fibrosis based on the PLCO dataset~\cite{PLCO}. The baseline (horizontal dashed lines) is determined using the method from~\cite{Guendel2018} (working point at max. average of per-class F1 scores). Decision threshold for our method is fixed at 0.5. \label{fig:samplereject}}
\end{figure}

\textbf{Uncertainty-driven Sample Rejection:}\, Given a model trained for the assessment of an arbitrary finding, one can directly estimate the prediction uncertainty $\hat{u} = 2/(\alpha + \beta) \in [0, 1]$. This is an orthogonal measure to the predicted probability, with increased values on out-of-distribution cases under the given model. One can use this measure for sample rejection, i.e., set a threshold $u_t$ and steer the system to not output its prediction on all cases with an expected uncertainty larger than $u_t$. Instead, these cases are labeled with the message \textit{"Do not know for sure; process case manually"}. In practice this leads to a significant increase in accuracy compared to the state-of-the-art on the remaining cases, as reported in Table~\ref{tab:results} and Figure~\ref{fig:samplereject}. For example, for the identification of granuloma, a rejection rate of 25\% leads to an increase of over 20\% in the micro-average F1 score. On the same abnormality, a 50\% rejection rate leads to a F1 score over 0.99 for the prediction of negative cases. We found no significant difference in average performance when using ensembles (see Figure~\ref{fig:samplereject}).\smallskip

\begin{table}[t]
\centering
\caption{Comparison between the reference method~\cite{Guendel2018} and several versions of our method calibrated at sample rejection rates of 0\%, 10\%, 25\% and 50\% (based on the PLCO dataset~\cite{PLCO}). Lesion refers to lesions of the bones or soft tissue.\label{tab:results}}
\begin{tabular}{L{1.8cm} C{2.5cm} C{1.7cm} C{1.9cm} C{1.9cm} C{1.8cm}}
&\multicolumn{5}{c}{\textbf{ROC-AUC}}\\
\cmidrule{2-6}
\textbf{Finding}&Guendel et al.~\cite{Guendel2018}&\textbf{Ours} [0\%]&\textbf{Ours} [10\%]&\textbf{Ours} [25\%]&\textbf{Ours} [50\%]\\
\midrule
Granuloma&0.83&0.85&0.87&\textbf{0.90}&\textbf{0.92}\\
Fibrosis&0.87&0.88&0.90&\textbf{0.92}&\textbf{0.94}\\
Scaring&0.82&0.81&0.84&\textbf{0.89}&\textbf{0.93}\\
Lesion&0.82&0.83&0.86&\textbf{0.88}&\textbf{0.90}\\
Cardiac Ab.&0.93&0.94&0.95&\textbf{0.96}&\textbf{0.97}\\
\midrule\midrule
\textbf{Average}&0.85&0.86&0.89&\textbf{0.91}&\textbf{0.93}\\
\bottomrule
\end{tabular}
\end{table}

\begin{figure}[t]
\centering
\includegraphics[height=3cm]{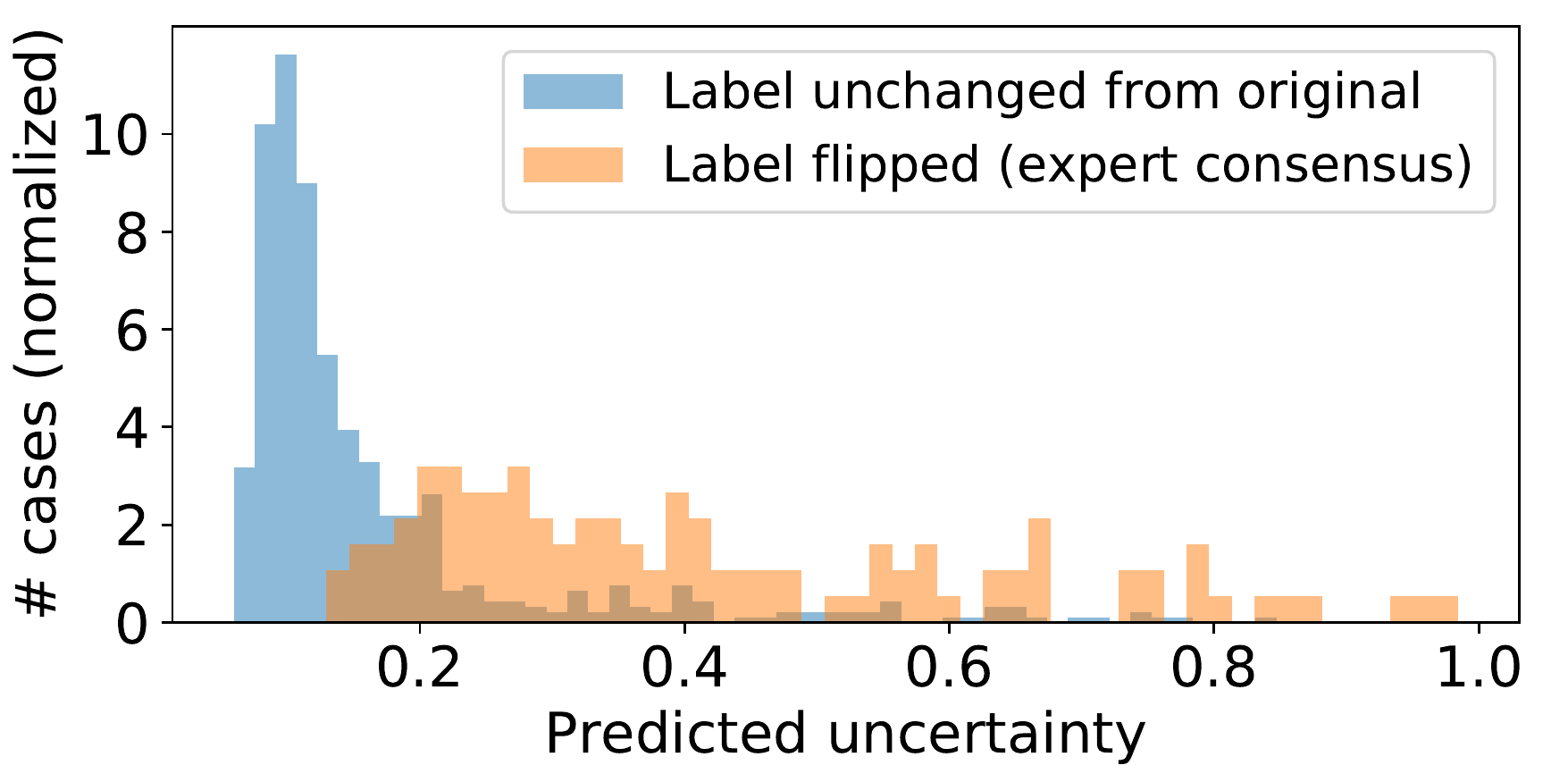}
\includegraphics[height=3cm]{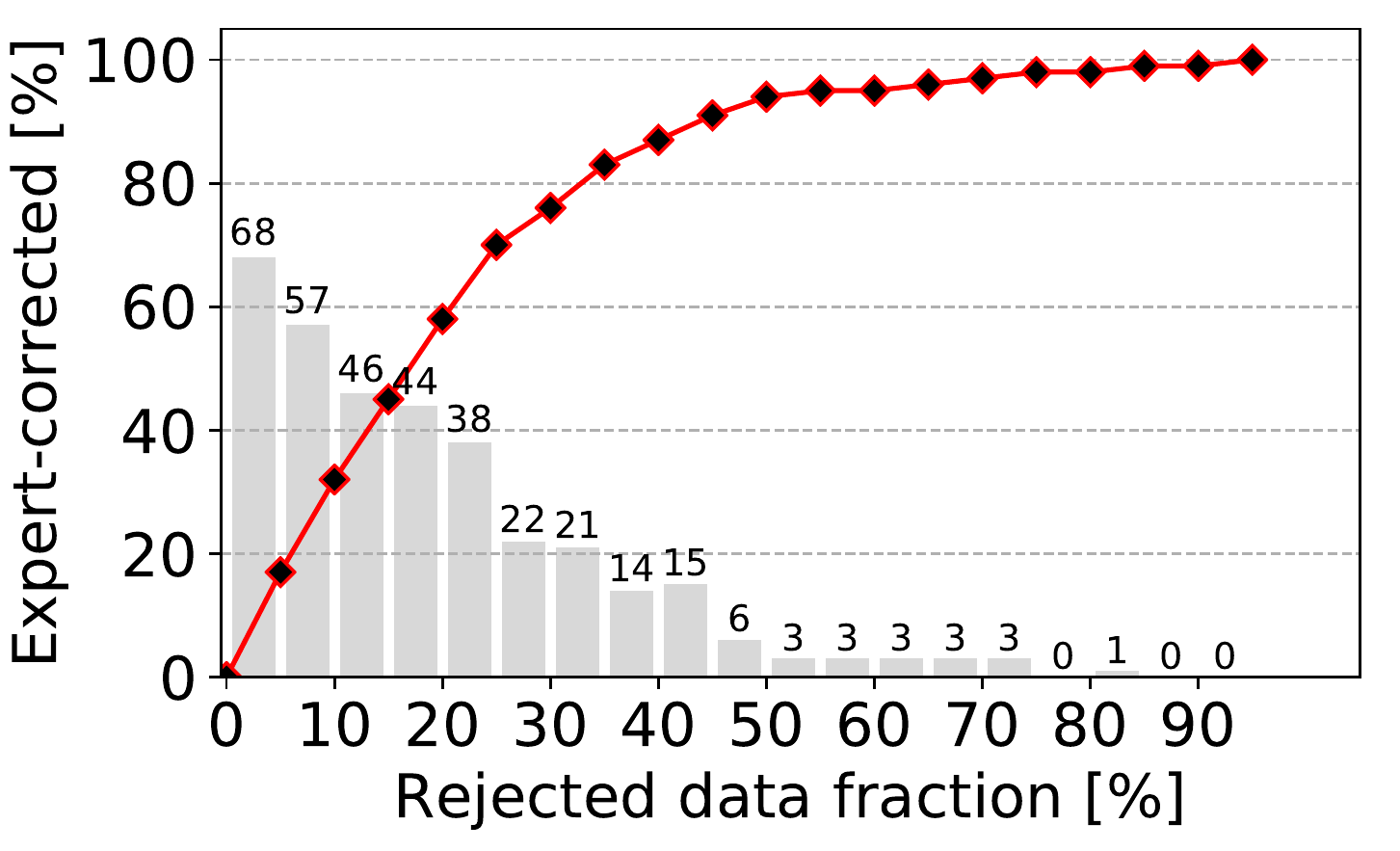}
\caption{\textbf{Left}: Predictive uncertainty distribution on 689 ChestX-Ray test images; a higher uncertainty is associated with cases of the critical set, which required a label correction according to expert committee. \textbf{Right}: Plot showing the capacity of the algorithm to eliminate cases from the critical set via sample rejection. Bars indicate the percentage of critical cases for each batch of 5\% rejected cases.\label{fig:corr}}
\end{figure}

\begin{figure*}[t]
\centering
\subfloat[$\hat{u},\hat{p}=0.90,0.45$]{
\includegraphics[height=2.8cm]{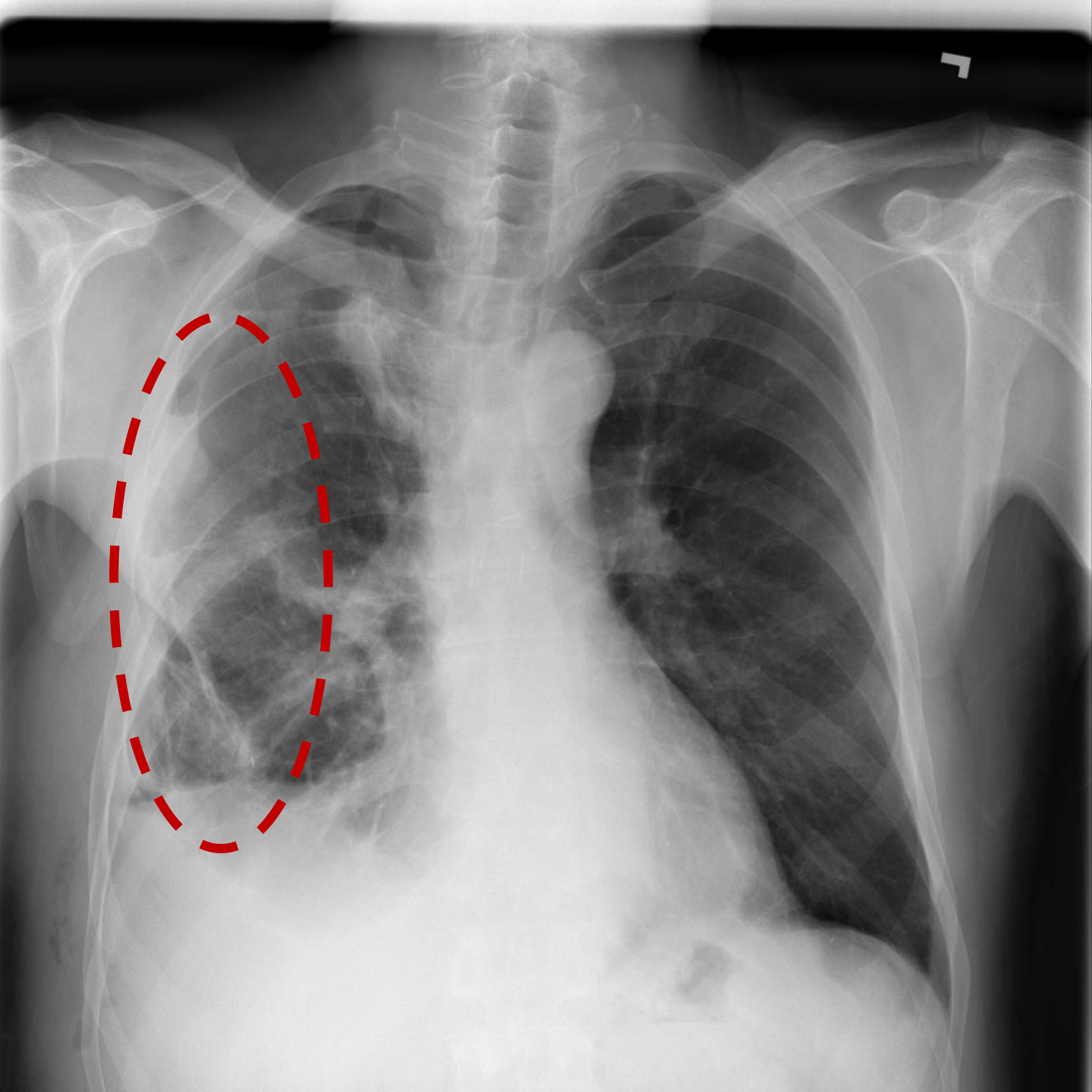}
\label{subfig:1}
}
\subfloat[$\hat{u},\hat{p}=0.93,0.48$]{
\includegraphics[height=2.8cm]{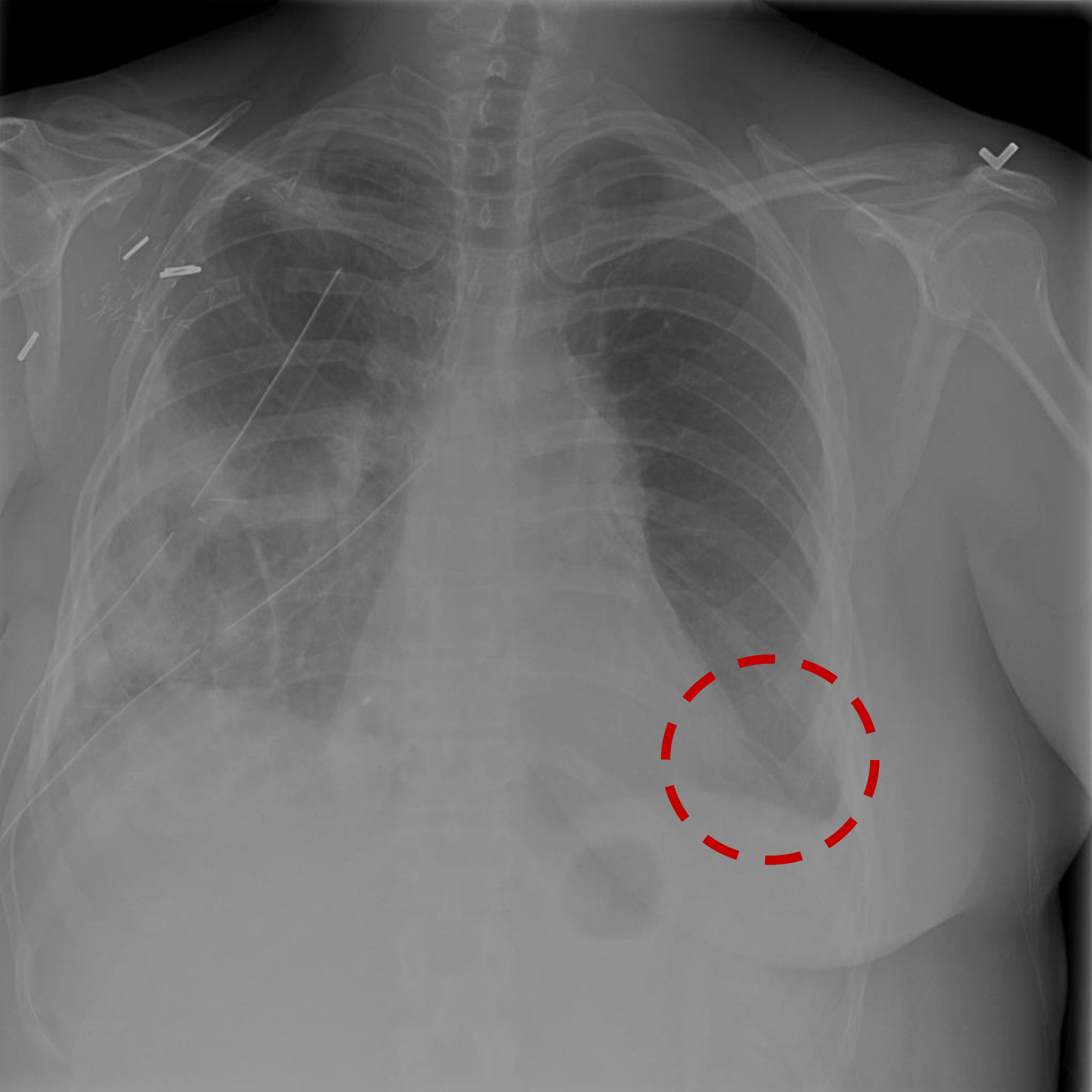}
\label{subfig:2}
}
\subfloat[$\hat{u},\hat{p}=0.54,0.65$]{
\includegraphics[height=2.8cm]{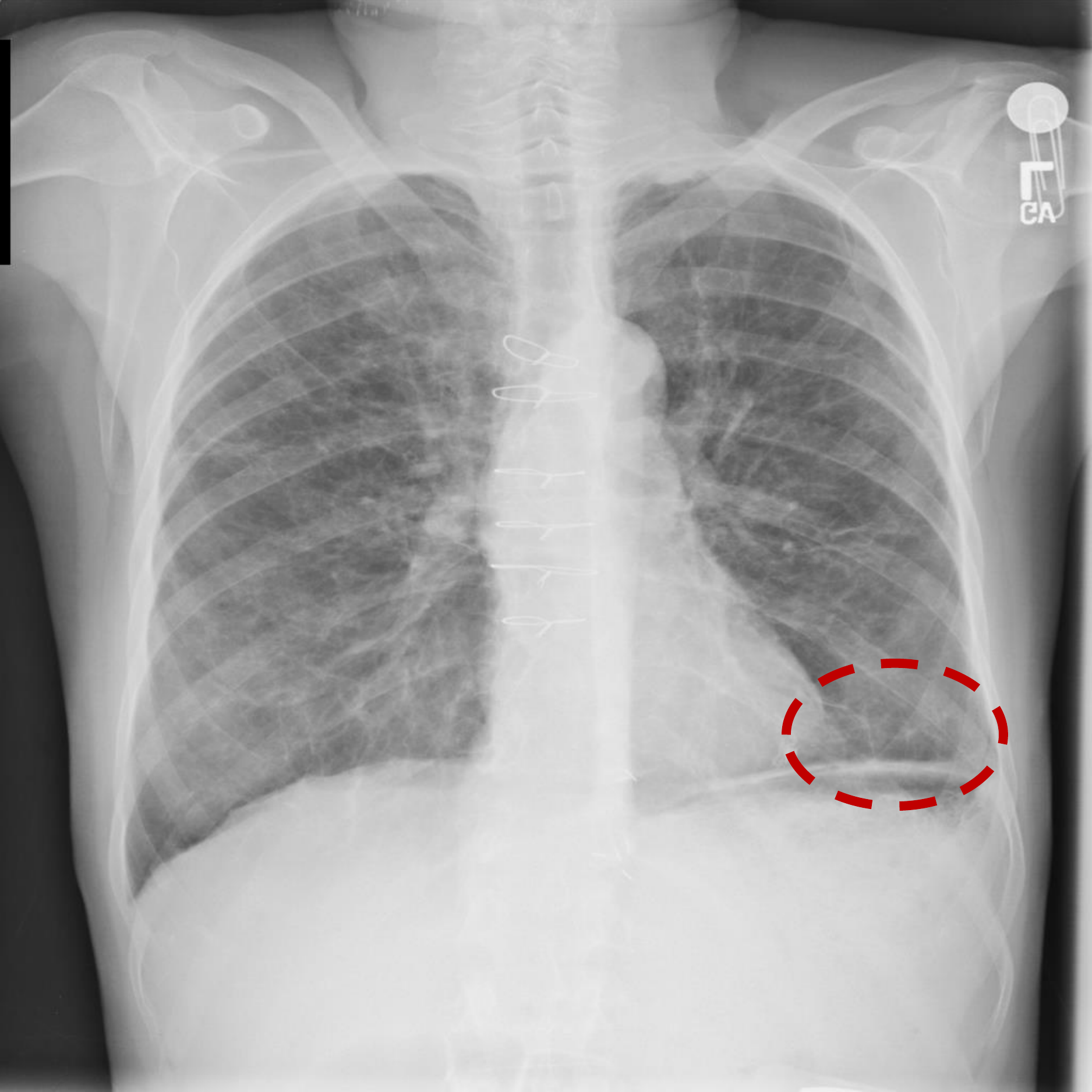}
\label{subfig:3}
}
\subfloat[$\hat{u},\hat{p}=0.11,0.05$]{
\includegraphics[height=2.8cm]{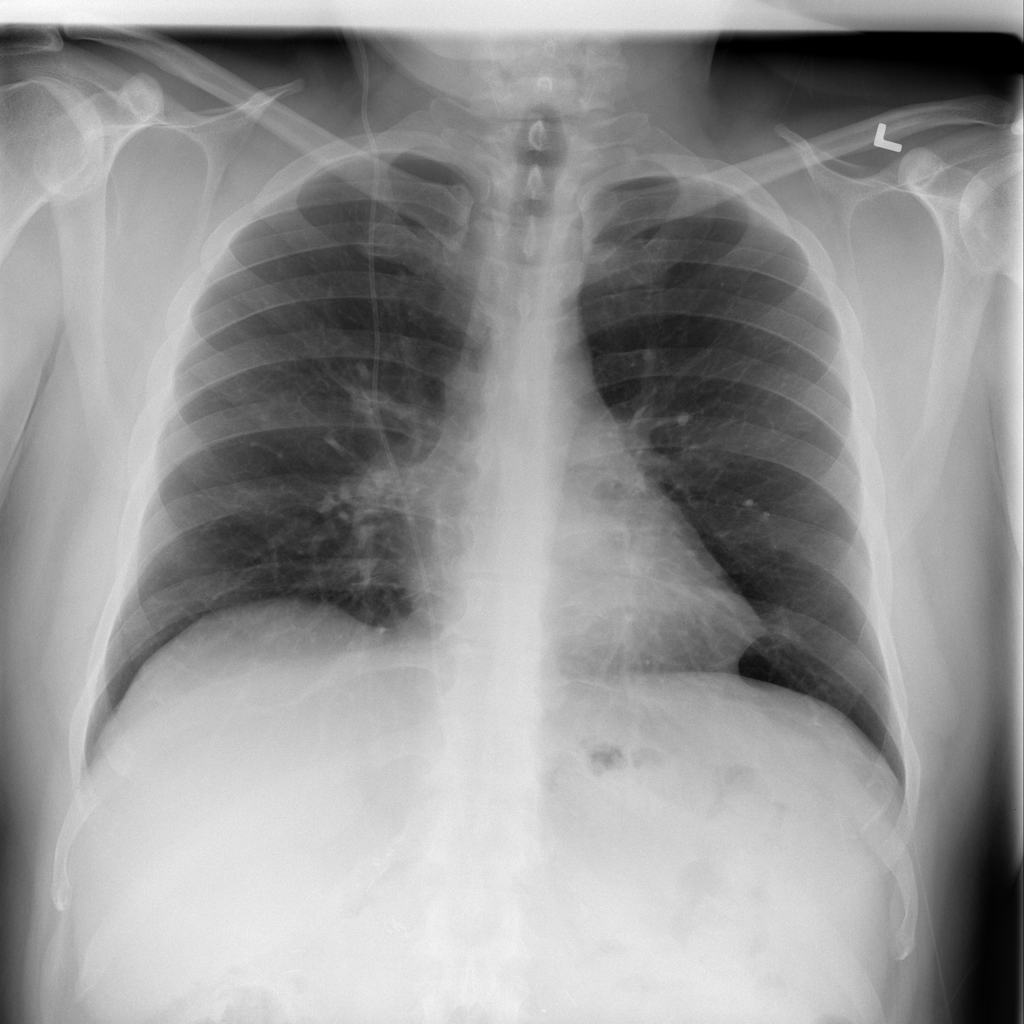}
\label{subfig:4}
}
\caption{ChestX-Ray8 test images assessed for pleural effusion ($\hat{u}$: est. uncertainty, $\hat{p}$: output probability; with affected regions circled in red). Figures~\ref{subfig:1}, \ref{subfig:2} and \ref{subfig:3} show positive cases of the critical set with high predictive uncertainty -- possibly explained by the atypical appearance of accumulated fluid in~\ref{subfig:1}, and poor quality of image~\ref{subfig:2}. Figure~\ref{subfig:4} shows a high confidence case with no pleural effusion.\label{fig:examples}}
\end{figure*}

\textbf{System versus Reader Uncertainty:}\, To provide an insight into the meaning of the uncertainty measure and its correlation with the difficulty of cases, we evaluated our system on the detection of pleural effusion (excess accumulation of fluid in the pleural cavity) based on the ChestX-Ray8 dataset. In particular, we analyzed the test set of 689 cases that were relabeled using an expert committee of 4 experts. We defined a so called \textit{critical set}, that contains only cases for which the label (positive or negative) was changed after the expert reevaluation. According to the committee, this set contained not only easy examples for which probably the NLP algorithm has failed to properly extract the correct labels from the radiographic report; but also difficult cases where either the image quality was limited or the evidence of effusion was very subtle. In Figure~\ref{fig:corr} (left), we empirically demonstrate that the uncertainty estimates of our algorithm correlate with the committee's decision to change the label. Specifically, for unchanged cases, our algorithm displayed very low uncertainty estimates (average 0.16) at an average AUC of 0.976 (rejection rate of 0\%). In contrast, on cases in the critical set, the algorithm showed higher uncertainties distributed between 0.1 and the maximum value of 1 (average 0.41). This empirically demonstrates the ability of the algorithm to recognize the cases where annotation errors occurred in the first place (through NLP or human reader error). In Figure~\ref{fig:corr} (right) we show how cases of the critical set can be effectively filtered out using sample rejection. Qualitative examples are shown in Figure~\ref{fig:examples}.\smallskip

\textbf{Uncertainty-driven Bootstrapping:}\, We also investigated the benefit of using bootstrapping based on the uncertainty measure on the example of plural effusion (ChestX-Ray8). 
We report performance as [\textit{AUC}; \textit{F1-score} (pos. class); \textit{F1-score} (neg. class)]. After training our method, the baseline performance was measured at $[0.89; 0.60; 0.92]$ on testing. We then eliminated 5\%, 10\% and 15\% of training samples with highest uncertainty, and retrained in each case on the remaining data. The metrics improved to $[0.90; 0.68; 0.92]_{5\%}$, $[0.91; 0.67; 0.94]_{10\%}$ and $[\textbf{0.93}; \textbf{0.69}; \textbf{0.94}]_{15\%}$ on the test set. This is a significant increase, demonstrating the potential of this strategy to improve the robustness of the model to the label noise. We are currently focused on further exploring this method.

\section{Conclusion}
In conclusion, this paper presents an effective method for the joint estimation of class probabilities and classification uncertainty in the context of chest radiograph assessment. Extensive experiments on two large datasets demonstrate a significant accuracy increase if sample rejection is performed based on the estimated uncertainty measure. In addition, we highlight the capacity of the system to distinguish radiographs with correct and incorrect labels according to a multi-radiologist-consensus user study, using the uncertainty measure only.\smallskip

The authors thank the National Cancer Institute for access to NCI's data collected by the Prostate, Lung, Colorectal and Ovarian (PLCO) Cancer Screening Trial. The statements contained herein are solely those of the authors and do not represent or imply concurrence or endorsement by NCI.

\ifx\anonymize\undefined
\textbf{Disclaimer} The concepts and information presented in this paper are based on research results that are not commercially available.
\fi

\bibliographystyle{splncs04}
\bibliography{main}

\end{document}